\documentclass[runningheads]{llncs}
\usepackage[T1]{fontenc}
\usepackage{array}
\usepackage{booktabs}

\usepackage{graphicx}
\begin{document}
\title{Explainable Automatic Grading with Neural Additive Models}
\author{Aubrey Condor\inst{1}\orcidID{0000-0002-9482-7003} \and
Zachary Pardos\inst{1}\orcidID{0000-0002-6016-7051}}
\authorrunning{A. Condor, Z. Pardos}
\institute{University of California Berkeley, Berkeley, CA 94720, USA
\email{aubrey\_condor@berkeley.edu}\\
}
\maketitle              % typeset the header of the contribution
\begin{abstract}
The use of automatic short answer grading (ASAG) models may help alleviate the time burden of grading while encouraging educators to frequently incorporate open-ended items in their curriculum. However, current state-of-the-art ASAG models are large neural networks (NN) often described as "black box", providing no explanation for which characteristics of an input are important for the produced output. This inexplicable nature can be frustrating to teachers and students when trying to interpret, or learn from an automatically-generated grade. To create a powerful yet intelligible ASAG model, we experiment with a type of model called a Neural Additive Model that combines the performance of a NN with the explainability of an additive model. We use a Knowledge Integration (KI) framework from the learning sciences to guide feature engineering to create inputs that reflect whether a student includes certain ideas in their response. We hypothesize that indicating the inclusion (or exclusion) of predefined ideas as features will be sufficient for the NAM to have good predictive power and interpretability, as this may guide a human scorer using a KI rubric. We compare the performance of the NAM with another explainable model, logistic regression, using the same features, and to a non-explainable neural model, DeBERTa, that does not require feature engineering.

\keywords{Explainable AI \and Automatic Grading \and ASAG \and Neural Additive Models.}
\end{abstract}
\section{Introduction}
It has been shown that the use of open-ended (OE) items is beneficial for student learning due to the generation effect \cite{bertsch2007generation} or in combination with self-explanation \cite{chi1994eliciting}. However, assessing OE items is time consuming for teachers \cite{hancock1995implementing} and consequently, educators default to using multiple choice (MC) questions instead. Automatic short answer grading (ASAG) may alleviate this time burden while encouraging educators to frequently incorporate OE items in their curriculum. 

Many of the best performing ASAG models include some variation of a deep neural network (NN) \cite{haller2022survey}. NNs are impressive predictors for high dimensional inputs like text embeddings, but predictive power tends to come at the cost of intelligibility. These models are often described as "black box" which means that users only have access to inputs and outputs, yet no information as to the process in-between. Further, unlike explainable models such as a logistic regression (LR), with NN ASAG models, we are not choosing which features the model may consider, and the model provides no explanation for which characteristics of an input are important for the produced output. Whereas educators typically use a scoring rubric to provide an explanation to stakeholders about which features of a given response substantiate the assigned score, and despite that attempts have been made to integrate scoring rubrics into ASAG models \cite{condor2022representing}, NN ASAG models provide no justification for their predictions. The inexplicable nature of ASAG models can be frustrating to both teachers and students when trying to make sense of, or learn from an automated grade. Teachers are unable to monitor student understandings at a fine-grained level, and students may not productively learn without knowing why they received a certain grade. It is critical that researchers find more explainable models if ASAG systems are to be of greater practical use.

In an effort to create a powerful yet intelligible ASAG model, we experiment with a type of model that combines the performance of a NN with the explainability of an additive model called a Neural Additive Model (NAM) \cite{agarwal2020neural}. NAMs allow us to visually examine the contribution of each feature to the final predicted score for each response, similar to testing the significance of a regression coefficient. For this type of model, we must engineer features of a response as inputs to the model instead of allowing the model to create its own text features like typical Large Language Model (LLM) classifiers do. The research questions we seek to answer include, (1) can NAMs provide intelligible automatic grading such that stakeholders can understand which features of a response are important for its prediction, and (2) is the predictive performance of NAMs better than that of legacy explainable models like a LR and commensurate with that of an LLM classifier? This research is unique as NAMs have not yet been explored for educational applications, or more specifically for explainable automatic grading. 

We demonstrate our NAM approach for ASAG with one item designed under a Knowledge Integration (KI) perspective from the learning sciences and corresponding rubric to guide our feature engineering. KI is a framework for strengthening science understanding that emphasizes incorporating new ideas and sorting out alternative perspectives with evidence \cite{linn2000designing}. We hypothesize that the inclusion (or exclusion) of predefined KI ideas as features will be sufficient for the NAM to have good predictive power, as this is precisely what would guide a human scorer using the KI rubric. We compare the performance of the NAM with another explainable model, a LR, using the same features as those used for the NAM, and to a non-explainable model, DeBERTa, an improved version of the popular BERT model \cite{he2021debertav3}. The DeBERTa model does not utilize the engineered features that are used by the NAM or LR model, but creates its own features from the text. For a more extensive comparison of the performance of these three models, we extend our analysis to include results from five additional questions from an open-source data set that has been used extensively in ASAG research - the Automatic Student Assessment Prize (ASAP) Short Answer Scoring (SAS) data \cite{prize2019hewlett}. 

\section{Related Work}
In this section, we describe previous works relating to Explainable AI and more particularly Explainable ASAG, as well as applications of NAMs.
\subsection{Explainable AI and ASAG}
A recent review article outlines the increase in demand for explainable AI (XAI), especially for people who are affected by AI driven decisions \cite{xu2019explainable}. They describe the rising popularity of NN models for their state-of-the-art  performance, despite that their inference processes are not known or interpretable. Many researchers aim to increase the explainability of AI systems. Initiatives such as DARPA’s "Explainable AI (XAI) program" \cite{gunning2019darpa} and the European Union’s "General Data Protection Regulation" which demands citizen’s rights to an explanation for decisions made by AI \cite{Goodman2017} are contributing to an increased demand for XAI. 
	
Only recently are researchers beginning to think about XAI in terms of ASAG models. Schlippe et al. \cite{schlippe2022explainability} surveyed over 70 educators about preferences for explainability in ASAG and learned that most prefer to see matches between student answers and exemplary answers to justify scores over other explainability methods like highlighting the importance of certain words. Poulton et al. \cite{poulton2021explaining} proposed an XAI framework using SHAP values to assess popular LLMs for ASAG such as BERT and RoBERTa. Zeng et al. \cite{zeng2022deep} investigated whether autograding models align with human graders in terms of the important words they use when assigning a grade by conducting a randomized controlled trial to see if highlighting words deemed important by an autograder can assist human grading. Singh et al. \cite{singh2023explaining} introduced Summarize and Score (SASC) to explain ‘text modules’, which map text to scalar values within LLMs, providing a natural language explanation of the module’s selectivity. Finally, Tornqvist et al. \cite{tornqvist2023exasag} introduced ExASAG, an explainable framework for ASAG that generates natural language explanations for predictions, and Condor and Pardos train a reinforcement learning agent to alter students' short responses so that these alterations provide insight as to why a ASAG score was assigned \cite{condor2022deep}.

\subsection{Applications of Neural Additive Models}
Although Neural Additive Models (NAMs) were introduced only a few years ago \cite{agarwal2020neural}, there has been substantial interest in their use as the machine learning community has increased efforts to promote understandable models. Some researchers have proposed altered or improved versions of NAMs for specific applications. Mariotti et al. \cite{mariotti2023exploring} explored the tension between interpretability and performance of NAMs and introduce a constrainable NAM (CNAM) which includes specifications for model regularization. The CNAM model is able to consider the performance-interpretability tradeoff during training, and is demonstrated to work well for both regression and classification tasks \cite{mariotti2023exploring}. Luber et al. \cite{luber2023structural} introduced Structural NAMs (SNAMs) which incorporate neural splines to a typical NAM. SNAMs offer improved intelligibility over NAMs by enabling direct interpretation of estimated parameters, as well as quantification of parameter uncertainty and post-hoc analysisr. Bouchiat et al. \cite{bouchiat2023laplace} take a Bayesian perspective and develop a Laplace approximated NAM (LA-NAM) to enhance interpretability, and Jo and Kim \cite{jo2022neural} introduced NAMs for multivariate time series modeling called NAM nowcasting (NAM-NC). Others have used NAMs for applications in finance \cite{chen2022monotonic}, mortality prediction \cite{moslehi2022interpretable}, and survival analysis \cite{utkin2022extension}. Importantly, we found no previous works that used NAMs for educational applications.

\section{Background}
In this section, we provide a description of the data, and a brief overview of NAMs, LR, and the DeBERTa model. 
\subsection{The Data}
\begin{figure}
\centering
  \includegraphics[width=10cm]{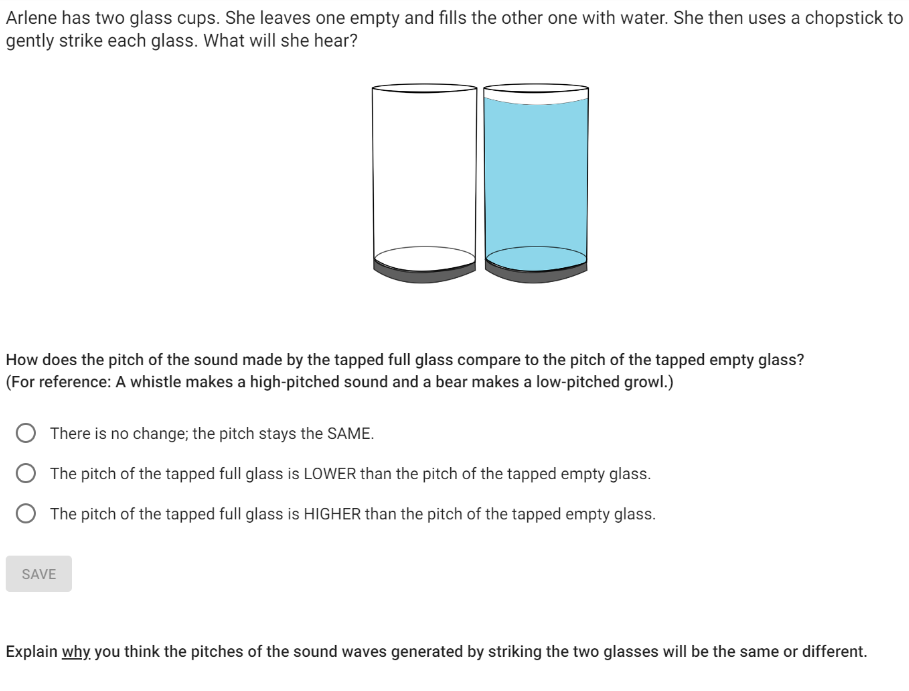}
  \caption{The Sound Waves item bundle}
  \label{soudwavesitem}
\end{figure}

\begin{figure}
\centering
  \includegraphics[width=10cm]{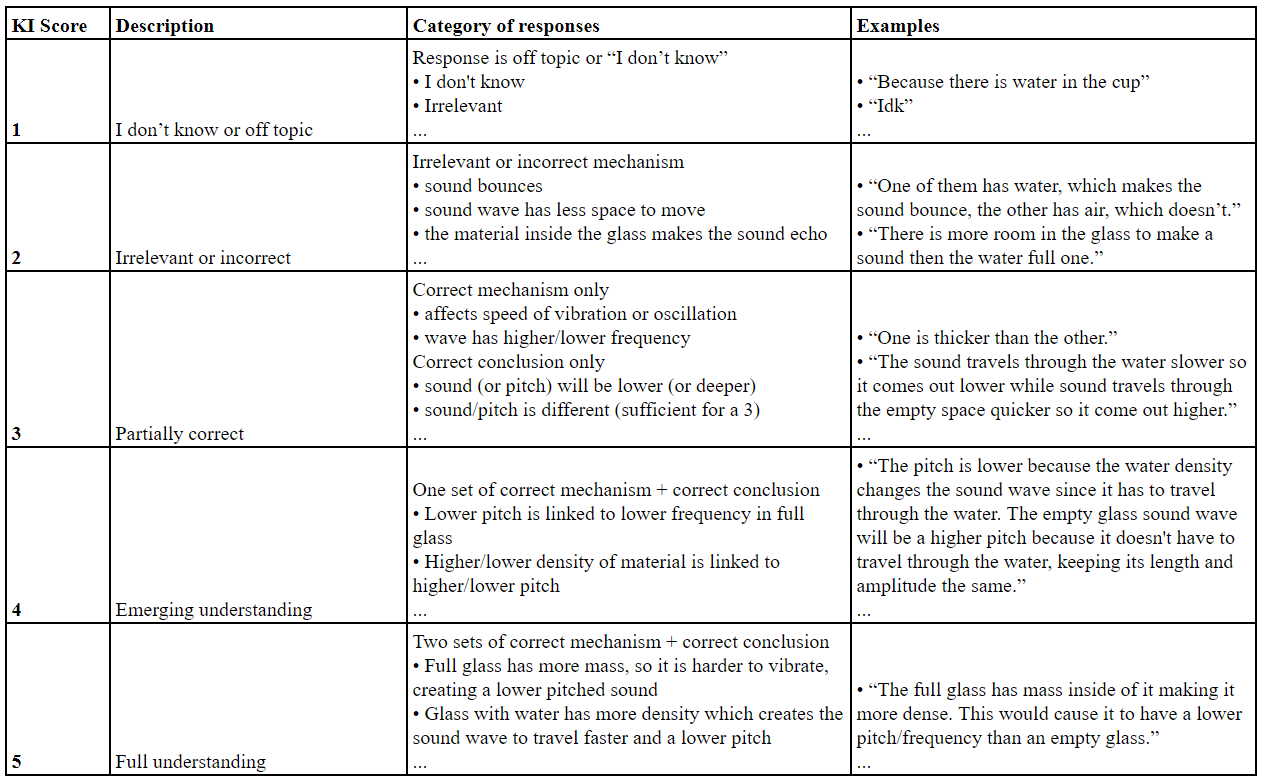}
  \caption{The KI Scoring Rubric}
  \label{soudwavesrubric}
\end{figure}

The KI item used for this project was collected during a previous research project at the Web-based Inquiry Science Environment (WISE) research center at the University of California, Berkeley consisting of OE science items designed for middle school students \cite{riordan2020empirical}. Students accessed the items via an online classroom system, and responses were scored with a Knowledge Integration (KI) rubric. For this project, we use one item from a unit about the physics of sound waves that engages students to refine their ideas about concepts such as wavelengths, frequency and pitch. For this item, students must distinguish how the pitch of sound made by tapping a full glass of water compares to the pitch made by tapping an empty glass. They are asked to explain why they think the pitch of the sound waves may be the same or different for the two glasses \cite{riordan2020empirical}. See Figure \ref{soudwavesitem} for a visual of the Soundwaves item. The KI data include 1,313 OE student responses that were carefully rated from 1 to 5 by subject matter experts. A detailed rubric was created by researchers and used for scoring which includes a description of the rating level, examples of correct/incorrect mechanisms and conclusions, and exemplar student responses that would fall into each category. \ref{soudwavesrubric} provides a sample of the KI scoring rubric.

The ASAP data used to extend our model comparison is from a 2012 Kaggle competition sponsored by the Hewlett Foundation, and consists of almost 13,000 short answer responses to 10 science and English questions \cite{prize2019hewlett}. We used only the 5 science questions to match the domain of the KI data. Each of the five items have between 1300-1800 OE student responses and corresponding human-rated scores between 1 and 4 with 4 being the most correct score. Each item has an expert created scoring rubric, and we aggregate the 5 items for training, testing and the resulting model comparisons.

\subsection{Neural Additive Models}
Neural Additive Models (NAMs) impose a restriction on the structure of NNs in order to make the model interpretable. NAMs belong to a family of models called Generalized Additive Models (GAMs) which have the form: 

\[g(E[y]) =  + f_1(x_1) +f_2(x_2) + ... + f_k(x_k)\]
Where \(x = (x_1, x_2, \dots, x_k)\) represents the input with K features and each \(f_i\) is a univariate shape function with \(E[f_i] = 0\) \cite{agarwal2020neural}. NAMs learn a linear combination of jointly-trained NNs where each NN attends to a single input feature. They can approximate complex, high-dimensional functions and their predictions are easily interpretable. NAMs provide advantages over standard GAMs such as their superior scalability with the use of GPU/TPU hardware developed for NNs, differentiability, and visual interpretability \cite{agarwal2020neural}. 

\subsection{Logistic Regression}
Multinomial LR is a classification model that predicts probabilities of different outcomes for a categorical dependent variable. To generalize to a K-class setting, the model runs K-1 independent binary LR models where one outcome is chosen as a “pivot” and other K-1 outcomes are separately regressed against the pivot outcome. We use the Limited-memory Broyden-Fletcher-Goldfarb-Shannon (LBfGS) algorithm for optimization \cite{fletcher2000practical}, and incorporate L2 regularization. 

\subsection{DeBERTa}
DeBERTa (Decoding-enhanced BERT with disentangled attention) is a Large Language Model (LLM) that improves upon the popular BERT \cite{devlin2018bert} model by using a different version of the standard attention mechanism \cite{vaswani2017attention} and a novel position encoding method. DeBERTa uses a ‘disentangled attention’ mechanism where each word is represented with two vectors - one to encode the word’s context, and another to encode the word’s relative position. Further, an enhanced masked decoder incorporates absolute positions to predict masked tokens during training. Finally, DeBERTa uses a unique adversarial training method to fine-tune the model’s generalizability performance on downstream tasks. We use an improved yet smaller version of the original DeBERTa model called DeBERTaV3-base \cite{he2021debertav3}, consisting of 12 layers, a hidden size of 768, and 86 million parameters.

\section{Methods}
To fit a NAM, we utilize the python library, "nam" \cite{kayid2020nams}, created by the researchers who introduced NAMs for easy implementation of the model. We altered the package code to accommodate our multi-class classification setting and allow for our evaluation procedure and corresponding metrics. Further described in the "Evaluation" section below, we use a cross validation (CV) evaluation procedure for model comparison, so for each model class, we train a total of 10 models on a different sample of the data. We train each of the 10 NAMs for 120 epochs with a batch size of 64 and a learning rate of 0.002. An epoch is one complete pass of the training data through the model, batch size represents the number of samples used in a single forward and backward pass through the network, and the learning rate governs the pace at which the model updates its parameter estimates. Further, we use a regularization technique called dropout with a value of 0.15 to help avoid over fitting. Hyperparameters were chosen using a standard grid search where different combinations of values are tested on a held-out validation set, and those that result in the best performance are chosen. Each of the DeBERTa models were trained for six epochs using a batch size of eight, and a learning rate of 0.0007. The LR models were fit using the scikit-learn "LogisticRegression" python library. 

\subsection{Feature Engineering}
To create features from student response texts for the NAM and LR models, we utilize KI phrases from the rubric that represent both correct, and incorrect mechanisms and conclusions. We chose to use KI ideas to create features because these are exactly what a human identifies to assign a grade to a student response within the KI framework. Because students often express their ideas using a variety of words (i.e., the idea that \textit{"sound moves faster in air"} could be stated as \textit{"sound travels faster through air"}), we scan the responses for n-grams that are similar to the chosen KI phrases. An n-gram is a series of n consecutive words within a string of text. For example, 2-grams of the phrase \textit{"the dog ran"} would be \textit{"the dog"} and \textit{"dog ran"}. We use n-grams of size n=1 through 5 for each response as students can also express the same ideas in a different number of words. Features for the ASAP science questions were created in the same manner (and use the same number of features for each item), although we provide the feature phrases only from the KI data for demonstration.

To calculate the similarity between a given feature phrase and an n-gram from a student response, we first embed both the n-gram and the phrase using sentence-BERT. Sentence-BERT utilizes siamese and triplet network structures to create semantically meaningful sentence embeddings \cite{reimers2019sentence}. We then calculate the cosine similarity of the KI phrase and the n-gram. Cosine similarity is a metric used to measure the similarity of text using the cosine of the angle between the vector embeddings of the two texts being compared. Once the KI phrase has been compared with all respective n-grams in a response, we use the largest cosine similarity found as the feature for that phrase. Thus the feature representation for each phrase provides a measure of whether or not some form of the phrase is included in the student response. We use 34 phrases and 28 key words from the KI Sound Wave Rubric (shown in Figure \ref{soudwavesrubric}) for a total of 62 features. The same number of phrases and words were chosen for each ASAP item from their corresponding scoring rubrics. The KI phrases are shown below, and the individual words were chosen from the phrases: \\
\\
\textit{"I don’t know", "idk", "sound bounces", "water blocks sound", "water mutes ringing", "sound moves more in air", "sound moves less in water", "sound echoes", "sound sinks in water", "sound moves fast in air", "frequency is height of wave", "pitch higher in full glass", "the density is different",  "water is more dense", "water vibrates less", "affects vibration",  "pitch is lower in water", "pitch is different",  "higher frequency in air", "frequency is different", "sound moves faster in water", "water has more mass",  "mass is different", "sound is denser in water", "sound is slower in water", "amplitude is number of waves", "pitch lower in empty glass", "air is less dense", "empty glass vibrates more", "vibration is different", "pitch is higher in air", "lower frequency in water", "sound moves slower in air", "empty glass less mass"} 

\subsection{Evaluation}
We use a standard ASAG evaluation metric in our results table: Quadratic Weighted Cohen’s Kappa (QWK). QWK reports the agreeability between two scores beyond random chance - a more robust measure than accuracy. A 5x2 cross validation (CV) paired t-test was used to evaluate the statistical significance of the difference in models. The 5x2 CV paired t-test is based on five iterations of twofold cross-validation, and is presented in \cite{dietterich_1998} as the recommended approximate statistical test for whether one machine learning model outperforms another because of it’s more acceptable type I error, and stronger statistical power than other methods such as McNemar’s test, or a paired t-test based on 10-fold CV. We used the QWK metric for the 5x2 CV paired t-tests (with 5 degrees of freedom). We perform a t-test comparing the QWK metric for the NAM to the LR model and the DeBERTa model. The null hypothesis for each t-test states that the QWK metric for the NAM is no different than the QWK metric for LR and DeBERTa. Further, we report the magnitude of the difference in performance of the NAM to LR and DeBERTa using the QWK metric from the CV model fits with a Cohen's D effect size. Cohen's D is essentially a difference of means, scaled by a standard deviation value \cite{kelley2012effect}. A low value of Cohen's D which represents a small magnitude of difference is typically below 0.20, and a large value would be around 0.80 or higher. We use a pooled standard deviation to calculate Cohen's D, which represents a weighted combination of the standard deviations from each model. 

\section{Results}
Comparisons of each ASAG model's performance are presented in tables \ref{CV} and \ref{Avg}. In Table \ref{CV}, we present results of the 5x2 CV paired t-test using the QWK metric for both the KI data and the aggregate ASAP data. We compare the NAM against both LR and DeBERTa. For the KI data, the NAM performs better than LR at a significance level of 0.05 (t=2.1732, p=0.0409), and the NAM performs slightly worse than the DeBERTa model but not statistically at a significance level of 0.05 (t=-0.8410, p=0.2194). Further, for the ASAP data, the NAM does not perform significantly better than the LR at a significance level of 0.05 (t=1.691, p=0.0758) although the p-value is quite small. For the ASAP data, we also see that the NAM performs worse than the DeBERTa model but not at a statistically significant level of 0.05 (t=-1.741, p=0.071), although similarly, the p-value is quite small. As shown by the CV average metrics in Table \ref{Avg}, on average, DeBERTa performs better than the NAM and the NAM performs better than the LR across both data sets. Additionally, in Table \ref{CV} we present the Cohen's D effect sizes for the difference in performance from the NAM to the two other models. The magnitude of difference between the NAM and both models is quite large for the KI data, with effect sizes greater than 1.0 for each comparison. For the ASAP data, the effect size is moderate when comparing the NAM and LR.

\begin{table}
\centering
  \caption{5x2 Cross Validation Paired t-test (5 df) and Cohen's D Effect Sizes}
  \label{CV}
  \begin{tabular}{cccc}
    \toprule
    NAM versus: & t(5) & p-val & Cohen's D \\
    \midrule
     & \textbf{KI data} & \\
    \midrule    
    DeBERTa & -0.8410 & 0.2194 &  -1.111 \\
    Log Reg & 2.1732 & 0.0409 &  1.003 \\
        \midrule
     & \textbf{ASAP data} & \\
    \midrule    
    DeBERTa & -1.741 & 0.0711 &  -1.070 \\
    Log Reg & 1.691 & 0.0758 &  0.222 \\
  \bottomrule
\end{tabular}
\end{table}

\begin{figure}
  \includegraphics[width=\textwidth]{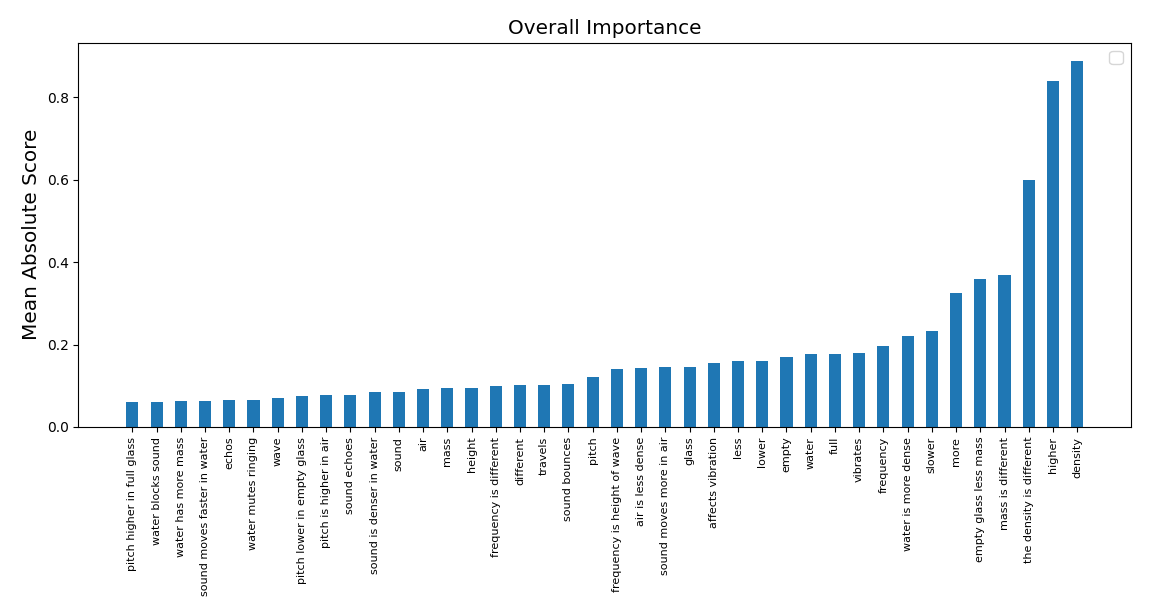}
  \caption{NAM Mean Feature Importance}
  \label{featImportance}
\end{figure}

\begin{figure}
  \includegraphics[width=\textwidth]{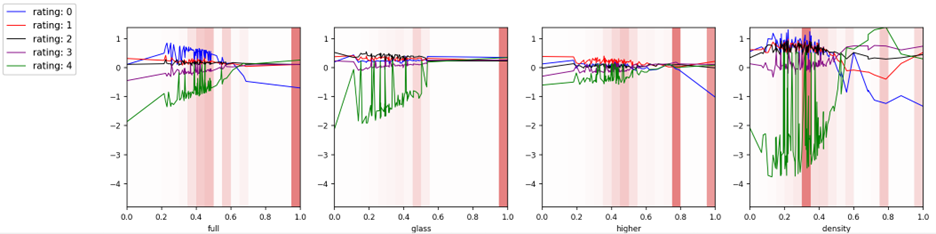}
  \caption{An Example of NAM shape functions with \textbf{all} rating categories}
  \label{shapeall}
\end{figure}

A visual of feature importance from the NAM for the KI data is presented in Figure \ref{featImportance}. We show the top 40 most important features in order of increasing importance. We see that the feature importance for the words \textit{"density"} and \textit{"higher"} is much larger than that of most other words and phrases. The phrase \textit{"the density is different"} also has a comparatively high importance. Further, in Figure \ref{density}, we present NAM Shape functions for the top 8 important words and phrases. As each feature of the NAM is handled independently by a learned shape function, we can see how the model makes its predictions, or in our case ratings, by graphing the shape functions for each feature. The scores are averaged and centered so that the shape functions can be directly compared on the same scale. As our NAM is a multi-class classifier, we can visualize five shape functions for each feature representing each of the five classes. We choose to show only the shape function for the lowest rating class and the highest rating class to retain the most important information yet avoid a messy, undecipherable visual. However, Figure \ref{shapeall} provides an example of shape plots with all five shape functions shown for reference. A negative y-axis score signals a low probability of a certain class, and a positive score represents a high probability. Noticeably, the shape functions are jagged-like with sharp jumps. \cite{agarwal2020neural} emphasizes the unique capability of NAMs to model highly “jumpy” 1-dimensional functions due to the use of an exponential-centered (ExU) non-linear function, whereas standard neural network models are bias towards smoothness with the use of Rectified Linear Unit (ReLU) functions. We could hypothesize that the true shape functions for the features in this project are more smooth, and in future work we could employ regularization techniques such as weight decay to provide smoother shape functions. On the same plot, we see the data density in the form of pink shaded bars. The darker the shade of pink, the more data there is in that region. In some areas with a really low data density, the model may not have had enough data to adequately learn the shape function.

\begin{table*}
\centering
  \caption{5x2 Cross Validation QWK Averages}
  \label{Avg}
  \begin{tabular}{ccc}
    \toprule
    Model & KI data & ASAP data \\
    \midrule
    DeBERTa & 0.7475 & 0.7176 \\
    Log Reg & 0.6929 & 0.6342 \\
    NAM & 0.7174 & 0.6473 \\
    \bottomrule
  \end{tabular}
\end{table*}

\section{Discussion}
The CV averages in Table \ref{Avg} give us an idea for how well each model performs for ASAG. We are not surprised to see that the NAM provides more predictive power than the LR, but not as much as the DeBERTa model. The DeBERTa model uses the entire student response as input text and learns many hundreds of features for its predictions, whereas with the NAM and LR, we are limited to the particular features we choose and it would be an unreasonable task to engineer the same number of features as the DeBERTa model learns. Additionally, the results of the 5x2 CV paired t-test in Table \ref{CV} give evidence that the NAM outperforms the LR model - across both data sets at a significance level of 0.10 and more notably with the KI data at a significance level of 0.05. Importantly, the NAM and the LR model use the same exact features. Further, the Cohen's D effect size values show that the magnitude of difference in performance of the models is quite large for the KI data, and smaller but still notable for the ASAP data. Not only is the difference statistically significant, but it seems to be practically meaningful. 

\begin{figure}
  \includegraphics[width=\textwidth]{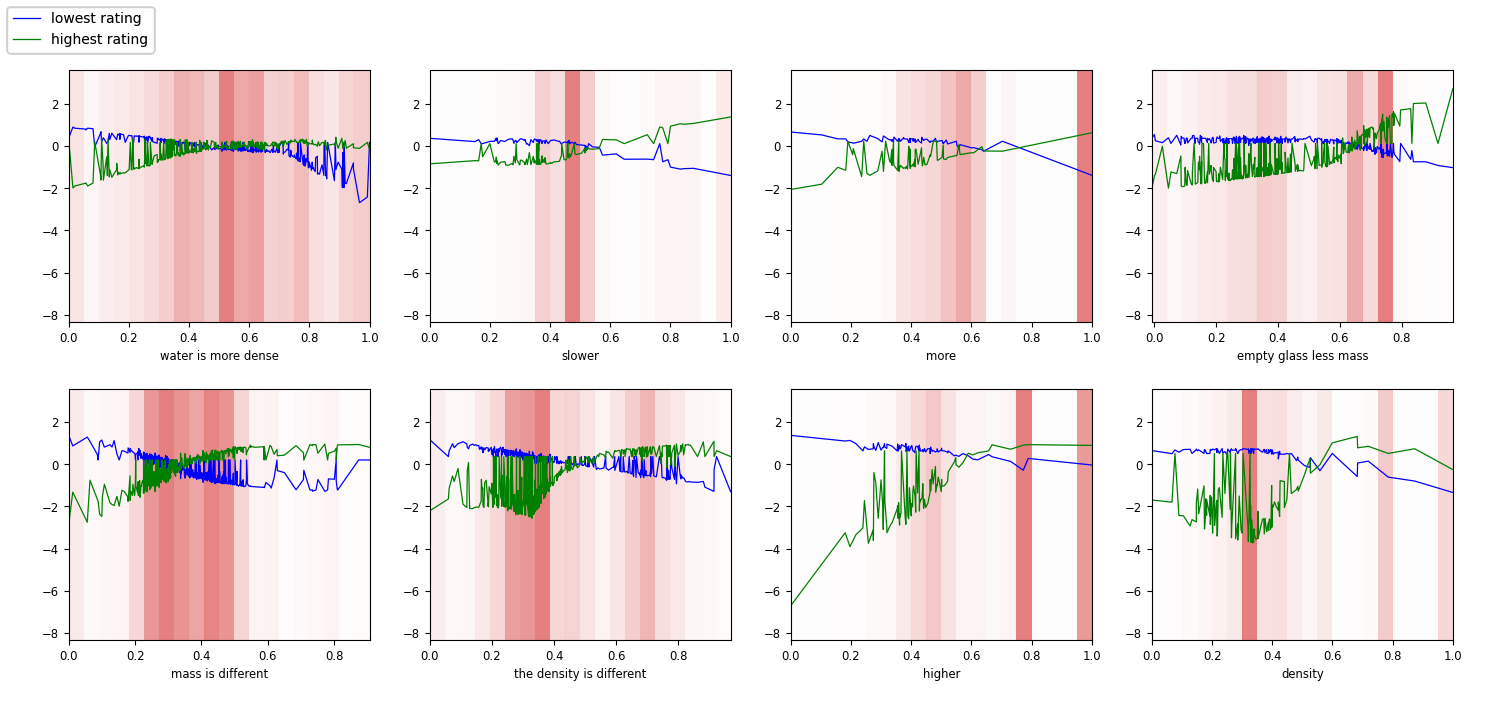}
  \caption{NAM shape functions of the highest and lowest rating category for the top 8 most important phrases/words. The y-axes show the log odds of predicting a given rating category, and the x-axes represent the range of similarity scores. The pink shades represent the data density at varying similarity scores.}
  \label{density}
\end{figure}

Figure \ref{featImportance} allows us to visually interpret the ASAG NAM through overall feature importance, and Figure \ref{density} gives us more detail about how each feature contributes with NAM shape functions. Human grading of OE responses can often consist of identifying which ideas - correct or incorrect - are included in a student's response. The KI scoring rubric allowed us to identify \textit{which} ideas (phrases and words) may help guide the NAM to correctly rate student responses, and the resulting NAM visualizations enable us to discern \textit{how} these ideas contribute to an automated rating. The feature importance plot can help stakeholders identify which ideas are most indicative of the student's grade, and the NAM shape function gives a notion of how the directionality of the model's prediction changes for different values of different rating classes. For example, for the \textit{“higher”} feature, the probability for the highest rating class (shown by the green line in Figure \ref{density}) goes down significantly with decreasing cosine similarities. So, including the word \textit{“higher”} in a response seems to correspond to higher scores. Additionally, with the density shading in the NAM shape function plots, we can infer the density of responses that include the given word or phrase. For example, the words \textit{“more”}, and \textit{“higher”} both have dense regions near high cosine similarities, so we can assume that many responses include these words or similar words. This can give an educator insight into which ideas students are more or less likely to write about.  

Limitations of using a NAM for ASAG include that the performance of the model in terms of its match to human ratings is less than that of a LLM ASAG model. Also, engineering features for the NAM is more costly than preparing student responses for a LLM ASAG model, as the model will create features from the text itself. Further, we only show results for science questions so we cannot conclude that our results would generalize to other short answer items from different subject areas. Our results suggest that the NAM may be a sufficient alternative to legacy explainable models, like a LR. Not only did the NAM generally perform better than the LR with the same features, but the NAM provides easily interpretable visualizations of the model’s prediction functions which can give an educator insights about their students’ understanding of the item content. Many researchers in the field of Learning Analytics still use LR over NNs for classification when the advantages of intelligibility outweigh that of performance \cite{le2018communication}\cite{deho2022existing}\cite{alonso2020predicting}\cite{misiejuk2021using}. The Learning Analytics community may benefit from investigating the use of NAMs for various classification tasks as well. In future work, we hope to experiment with using NAMs for ASAG with different question types from different domains, and perform qualitative interviews with educators to see if the NAM visualizations are understandable and helpful. 

\bibliographystyle{splncs04}
\bibliography{NAM}

\begin{thebibliography}{10}
\providecommand{\url}[1]{\texttt{#1}}
\providecommand{\urlprefix}{URL }
\providecommand{\doi}[1]{https://doi.org/#1}

\bibitem{agarwal2020neural}
Agarwal, R., Frosst, N., Zhang, X., Caruana, R., Hinton, G.E.: Neural additive models: Interpretable machine learning with neural nets. arXiv preprint arXiv:2004.13912  (2020)

\bibitem{alonso2020predicting}
Alonso-Fern{\'a}ndez, C., Mart{\'\i}nez-Ortiz, I., Caballero, R., Freire, M., Fern{\'a}ndez-Manj{\'o}n, B.: Predicting students' knowledge after playing a serious game based on learning analytics data: A case study. Journal of Computer Assisted Learning  \textbf{36}(3),  350--358 (2020)

\bibitem{bertsch2007generation}
Bertsch, S., Pesta, B.J., Wiscott, R., McDaniel, M.A.: The generation effect: A meta-analytic review. Memory \& cognition  \textbf{35}(2),  201--210 (2007)

\bibitem{bouchiat2023laplace}
Bouchiat, K., Immer, A., Y{\`e}che, H., R{\"a}tsch, G., Fortuin, V.: Laplace-approximated neural additive models: Improving interpretability with bayesian inference. arXiv preprint arXiv:2305.16905  (2023)

\bibitem{chen2022monotonic}
Chen, D., Ye, W.: Monotonic neural additive models: Pursuing regulated machine learning models for credit scoring. In: Proceedings of the Third ACM International Conference on AI in Finance. pp. 70--78 (2022)

\bibitem{chi1994eliciting}
Chi, M.T., De~Leeuw, N., Chiu, M.H., LaVancher, C.: Eliciting self-explanations improves understanding. Cognitive science  \textbf{18}(3),  439--477 (1994)

\bibitem{condor2022deep}
Condor, A., Pardos, Z.: A deep reinforcement learning approach to automatic formative feedback. International Educational Data Mining Society  (2022)

\bibitem{condor2022representing}
Condor, A., Pardos, Z., Linn, M.: Representing scoring rubrics as graphs for automatic short answer grading. In: International Conference on Artificial Intelligence in Education. pp. 354--365. Springer (2022)

\bibitem{deho2022existing}
Deho, O.B., Zhan, C., Li, J., Liu, J., Liu, L., Duy~Le, T.: How do the existing fairness metrics and unfairness mitigation algorithms contribute to ethical learning analytics? British Journal of Educational Technology  \textbf{53}(4),  822--843 (2022)

\bibitem{devlin2018bert}
Devlin, J., Chang, M.W., Lee, K., Toutanova, K.: Bert: Pre-training of deep bidirectional transformers for language understanding. arXiv preprint arXiv:1810.04805  (2018)

\bibitem{dietterich_1998}
Dietterich, T.G.: Approximate statistical tests for comparing supervised classification learning algorithms. Neural Computation  \textbf{10}(7),  1895–1923 (1998). \doi{10.1162/089976698300017197}

\bibitem{fletcher2000practical}
Fletcher, R.: Practical methods of optimization. John Wiley \& Sons (2000)

\bibitem{Goodman2017}
Goodman, B., Flaxman, S.: European union regulations on algorithmic decision making and a ”right to explanation”. AI magazine  \textbf{38}(2),  781--796 (2017)

\bibitem{gunning2019darpa}
Gunning, D., Aha, D.: Darpa’s explainable artificial intelligence (xai) program. AI magazine  \textbf{40}(2),  44--58 (2019)

\bibitem{haller2022survey}
Haller, S., Aldea, A., Seifert, C., Strisciuglio, N.: Survey on automated short answer grading with deep learning: from word embeddings to transformers. arXiv preprint arXiv:2204.03503  (2022)

\bibitem{hancock1995implementing}
Hancock, C.L.: Implementing the assessment standards for school mathematics: Enhancing mathematics learning with open-ended questions. The Mathematics Teacher  \textbf{88}(6),  496--499 (1995)

\bibitem{he2021debertav3}
He, P., Gao, J., Chen, W.: Debertav3: Improving deberta using electra-style pre-training with gradient-disentangled embedding sharing. arXiv preprint arXiv:2111.09543  (2021)

\bibitem{jo2022neural}
Jo, W., Kim, D.: Neural additive models for nowcasting. arXiv preprint arXiv:2205.10020  (2022)

\bibitem{kayid2020nams}
Kayid, A., Frosst, N., Hinton, G.E.: Neural additive models library (2020)

\bibitem{kelley2012effect}
Kelley, K., Preacher, K.J.: On effect size. Psychological methods  \textbf{17}(2), ~137 (2012)

\bibitem{le2018communication}
Le, C.V., Pardos, Z.A., Meyer, S.D., Thorp, R.: Communication at scale in a mooc using predictive engagement analytics. In: Artificial Intelligence in Education: 19th International Conference, AIED 2018, London, UK, June 27--30, 2018, Proceedings, Part I 19. pp. 239--252. Springer (2018)

\bibitem{linn2000designing}
Linn, M.C.: Designing the knowledge integration environment. International Journal of Science Education  \textbf{22}(8),  781--796 (2000)

\bibitem{luber2023structural}
Luber, M., Thielmann, A., S{\"a}fken, B.: Structural neural additive models: Enhanced interpretable machine learning. arXiv preprint arXiv:2302.09275  (2023)

\bibitem{mariotti2023exploring}
Mariotti, E., Moral, J.M.A., Gatt, A.: Exploring the balance between interpretability and performance with carefully designed constrainable neural additive models. Information Fusion p. 101882 (2023)

\bibitem{misiejuk2021using}
Misiejuk, K., Wasson, B., Egelandsdal, K.: Using learning analytics to understand student perceptions of peer feedback. Computers in human behavior  \textbf{117},  106658 (2021)

\bibitem{moslehi2022interpretable}
Moslehi, S., Mahjub, H., Farhadian, M., Soltanian, A.R., Mamani, M.: Interpretable generalized neural additive models for mortality prediction of covid-19 hospitalized patients in hamadan, iran. BMC Medical Research Methodology  \textbf{22}(1), ~339 (2022)

\bibitem{poulton2021explaining}
Poulton, A., Eliens, S.: Explaining transformer-based models for automatic short answer grading. In: Proceedings of the 5th International Conference on Digital Technology in Education. pp. 110--116 (2021)

\bibitem{prize2019hewlett}
Prize, A.S.A.: The hewlett foundation: Automated essay scoring (2019)

\bibitem{reimers2019sentence}
Reimers, N., Gurevych, I.: Sentence-bert: Sentence embeddings using siamese bert-networks. arXiv preprint arXiv:1908.10084  (2019)

\bibitem{riordan2020empirical}
Riordan, B., Bichler, S., Bradford, A., King~Chen, J., Wiley, K., Gerard, L., Linn, M.: An empirical investigation of neural methods for content scoring of science explanations. In: Proceedings of the fifteenth workshop on innovative use of NLP for building educational applications (2020)

\bibitem{schlippe2022explainability}
Schlippe, T., Stierstorfer, Q., Koppel, M.t., Libbrecht, P.: Explainability in automatic short answer grading. In: International Conference on Artificial Intelligence in Education Technology. pp. 69--87. Springer (2022)

\bibitem{singh2023explaining}
Singh, C., Hsu, A.R., Antonello, R., Jain, S., Huth, A.G., Yu, B., Gao, J.: Explaining black box text modules in natural language with language models. arXiv preprint arXiv:2305.09863  (2023)

\bibitem{tornqvist2023exasag}
Tornqvist, M., Mahamud, M., Guzman, E.M., Farazouli, A.: Exasag: Explainable framework for automatic short answer grading. In: Proceedings of the 18th Workshop on Innovative Use of NLP for Building Educational Applications (BEA 2023). pp. 361--371 (2023)

\bibitem{utkin2022extension}
Utkin, L., Konstantinov, A.: An extension of the neural additive model for uncertainty explanation of machine learning survival models. In: Cyber-Physical Systems: Intelligent Models and Algorithms, pp. 3--13. Springer (2022)

\bibitem{vaswani2017attention}
Vaswani, A., Shazeer, N., Parmar, N., Uszkoreit, J., Jones, L., Gomez, A.N., Kaiser, {\L}., Polosukhin, I.: Attention is all you need. Advances in neural information processing systems  \textbf{30} (2017)

\bibitem{xu2019explainable}
Xu, F., Uszkoreit, H., Du, Y., Fan, W., Zhao, D., Zhu, J.: Explainable ai: A brief survey on history, research areas, approaches and challenges. In: CCF international conference on natural language processing and Chinese computing. pp. 563--574. Springer (2019)

\bibitem{zeng2022deep}
Zeng, Z., Li, X., Gasevic, D., Chen, G.: Do deep neural nets display human-like attention in short answer scoring? In: Proceedings of the 2022 conference of the north American chapter of the Association for Computational Linguistics: Human language technologies. pp. 191--205 (2022)

\end{thebibliography}
\end{document}